\documentclass[lettersize,journal,compsoc]{IEEEtran}

\usepackage{graphicx, times, amsmath, amssymb, cite, subfigure, stfloats, subfigure,booktabs, mathtools,bm,bbm,multirow, epstopdf}

\usepackage{amsmath,amssymb,amsfonts}
\usepackage[ruled,norelsize]{algorithm2e}

\makeatletter
\newcommand{\removelatexerror}{\let\@latex@error\@gobble}
\makeatother

\begin{document}

\title{Cross-Task Inconsistency Based Active Learning (CTIAL) for Emotion Recognition}

\author{
    Yifan~Xu, Xue Jiang, and Dongrui~Wu
    \IEEEcompsocitemizethanks{
  	\IEEEcompsocthanksitem
        Y.~Xu, X.~Jiang and D.~Wu are with the Key Laboratory of the Ministry of Education for Image Processing and Intelligent Control, School of Artificial Intelligence and Automation, Huazhong University of Science and Technology, Wuhan 430074, China. Email: yfxu@hust.edu.cn, xuejiang@hust.edu.cn, drwu09@gmail.com.}
    }

\IEEEtitleabstractindextext{
\begin{abstract}
Emotion recognition is a critical component of affective computing. Training accurate machine learning models for emotion recognition typically requires a large amount of labeled data. Due to the subtleness and complexity of emotions, multiple evaluators are usually needed for each affective sample to obtain its ground-truth label, which is expensive. To save the labeling cost, this paper proposes an inconsistency-based active learning approach for cross-task transfer between emotion classification and estimation. Affective norms are utilized as prior knowledge to connect the label spaces of categorical and dimensional emotions. Then, the prediction inconsistency on the two tasks for the unlabeled samples is used to guide sample selection in active learning for the target task. Experiments on within-corpus and cross-corpus transfers demonstrated that cross-task inconsistency could be a very valuable metric in active learning. To our knowledge, this is the first work that utilizes prior knowledge on affective norms and data in a different task to facilitate active learning for a new task, even the two tasks are from different datasets.
\end{abstract}

}

\maketitle

\section{Introduction}\label{sec:intro}

\IEEEPARstart{E}motion recognition is an important part of affective computing, which focuses on identifying and understanding human emotions from facial expressions \cite{pantic2000}, body gestures \cite{Kaliouby2004}, speech \cite{wu2019affect}, physiological signals \cite{drwuPIEEE2023}, etc. It has potential applications in healthcare and human-machine interactions, e.g., emotion health surveillance \cite{Jiang2020} and emotion-based music recommendation \cite{Ayata2018}.

Emotions can be represented categorically (discretely) or dimensionally (continuously). A typical example of the former is Ekman's six basic emotions \cite{ekman1987universals}. Typical dimensional emotion representations include the pleasure-arousal-dominance model \cite{mehrabian1980basic} (pleasure is often replaced by its opposite, valence), and the circumplex model \cite{russell1980circumplex} that only considers valence and arousal. Both categorical emotion classification (CEC) and dimensional emotion estimation (DEE) are considered in this paper.

Accurate emotion recognition usually requires a large amount of labeled training data. However, labeling affective data is expensive because emotion is subtle and has large individual differences. Each affective sample needs to be evaluated by multiple annotators to obtain its `ground-truth' label. Active learning (AL)~\cite{Settles2009} and transfer learning (TL)~\cite{pan2010survey} are promising solutions to alleviate the labeling effort.

AL selects the most useful samples to query for their labels, improving the model performance for a limited annotation budget. The key is to define an appropriate measure of sample usefulness. For CEC from facial images, Muhammad and Alhamid \cite{muhammad2017user} adopted entropy as the uncertainty measure and selected samples with high entropy to annotate. Zhang \emph{et al.} \cite{zhang2015dynamic} used the distance to decision boundary as the uncertainty measure in AL for speech emotion classification. They selected samples with medium certainty to avoid noisy ones and allocated different numbers of annotators for each sample adaptively. For AL in DEE, Wu \emph{et al.} \cite{wu2019active} considered greedy sampling in both the feature space and the label space. Their approach was later extended to multi-task DEE \cite{wu2019affect}. Abdelwahab and Busso \cite{abdelwahab2019active} also verified the effectiveness of greedy sampling based AL approaches in valence and arousal estimation using deep neural networks.

TL utilizes knowledge from relevant (source) tasks to facilitate the learning for a new (target) task \cite{wu2020transfer,drwuNTL2023}. In task-homogeneous TL, i.e., the source and target domains have the same label space, the key is to reduce their distribution discrepancies \cite{li2021can}. For example, Zhou and Chen \cite{zhou2019transferable} employed class-wise adversarial domain adaptation to transfer from spontaneous emotional speeches of children to acted corpus of adults. A more challenging scenario is task-heterogenous TL, where the source and target domains have different label spaces. Pre-training on large datasets of relevant (but may not be exactly the same) tasks and then fine-tuning on task-specific data was used in facial emotion recognition \cite{kaya2017video,ngo2019facial, sugianto2019cross}. Zhao \emph{et al.} \cite{zhao2020speech} adopted age and gender prediction as source tasks, and supplemented the extracted features to target tasks of speech emotion classification and estimation. Park \emph{et al.} \cite{park2019toward} trained models on text corpora with categorical emotion labels to estimate their dimensional emotions. They ordered the discrete emotions along valence, arousal, and dominance dimensions according to domain knowledge \cite{Mohammad2018} and reduced the earth mover's distance to optimize the model.

\begin{figure*}[htbp]\centering
\includegraphics[width=0.85\linewidth,clip]{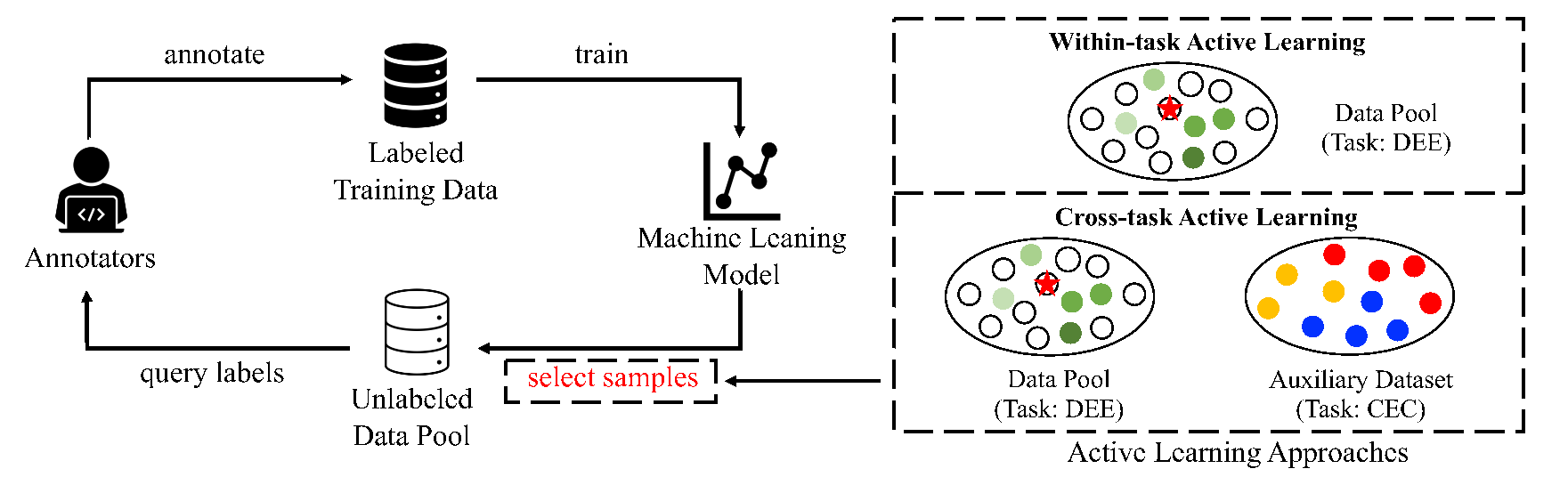}
\caption{Within-task AL for DEE, and CTIAL for cross-task transfer from CEC to DEE.} \label{fig:al_flowchart}
\end{figure*}

This paper proposes cross-task inconsistency based active transfer learning (CTIAL) between CEC and DEE, which has not been explored before. We aim to reduce the labeling efforts in the task-heterogeneous TL scenario where a homogenous source dataset suitable for transfer is hard to obtain, but a heterogeneous one is available. CTIAL enhances the efficiency of sample selection in AL for the target task, by exploiting the source task knowledge. Fig.~\ref{fig:al_flowchart} illustrates the difference between cross-task AL and the traditional within-task AL.

To transfer knowledge between CEC and DEE tasks, we first train models for CEC and DEE separately using the corresponding labeled data and obtain their predictions on the unlabeled samples. Affective norms, normative emotional ratings for words~\cite{bradley1999affective}, are utilized as domain knowledge to map the estimated categorical emotion probabilities into the dimensional emotion space. A cross-task inconsistency (CTI) is computed in this common label space and used as an informativeness measure in AL. By further integrating the CTI with other metrics, e.g., uncertainty in CEC or diversity in DEE, we can identify the most useful unlabeled samples to annotate for the target task and merge them into the labeled dataset to update the model.

Our contributions are:
\begin{enumerate}
	\item We propose CTI to measure the prediction inconsistency between CEC and DEE tasks for cross-task AL.
	\item We integrate CTI with other metrics in within-task AL to further improve the sample query efficiency.
	\item Within- and cross-corpus experiments demonstrated the effectiveness of CTIAL in cross-task transfers.
\end{enumerate}
To our knowledge, this is the first work that utilizes prior knowledge on affective norms and data in a different task to facilitate AL for a new task, even though the two tasks are from different datasets.

The remainder of the paper is organized as follows: Section~\ref{sec:method} introduces the proposed CTIAL approach. Section~\ref{sec:setup} describes the datasets and the experimental setup. Sections~\ref{sec:exp_CEC} and \ref{sec:exp_DEE} present experiment results on cross-task transfers from DEE to CEC, and from CEC to DEE, respectively. Section~\ref{sec:conclusion} draws conclusions.

\section{Methodology} \label{sec:method}

This section introduces the CTI measure and its application in cross-task AL.

\subsection{Problem Setting}

Consider the transfer between CEC and DEE, i.e., one is the source task, and the other is the target task. The source task has a large amount of labeled samples, whereas the target task has only a few. AL selects the most useful target samples from the unlabeled data pool and queries for their labels.

Denote the dataset with categorical and dimensional emotion annotations as $\mathcal{D}^\text{Cat}=\{(\bm{x}_i, \bm{y}_i^\text{Cat})\}_{i=1}^{N_\text{Cat}}$ and $\mathcal{D}^\text{Dim}=\{(\bm{x}_i, \bm{y}_i^\text{Dim})\}_{i=1}^{N_\text{Dim}}$, respectively, where $\bm{x}_i$ in $\mathcal{D}^\text{Cat}$ and $\mathcal{D}^\text{Dim}$ are from the same feature space, $\bm{y}_i^\text{Cat}\in \mathcal{R}^{|E|}$ is the one-hot encoding label of the emotion category set $E$, $\bm{y}_i^\text{Dim}\in \mathcal{R}^{|D|}$ is the dimensional emotion label of the dimension set $D$. An example of the emotion sets is $E$=\{angry, happy, sad, neutral\} and $D$=\{valence, arousal, dominance\}. The unlabeled data pool $P=\{\bm{x}_i\}_{i=1}^{N_P}$ has homogenous features with $\mathcal{D}^\text{Cat}$ and $\mathcal{D}^\text{Dim}$. The target dataset consists of $P$ and a few labeled samples.

\subsection{Expert Knowledge}\label{subsec:dk}

Affective norms, i.e., dimensional emotion ratings for words~\cite{bradley1999affective,Mohammad2018}, can be utilized as domain knowledge to establish the connection between the label spaces of categorical and dimensional emotions~\cite{zhou2020fine, park2019toward}. This paper uses the NRC Valence-Arousal-Dominance Lexicon \cite{Mohammad2018}, where 20,000 English words were manually annotated with valence, arousal and dominance scores via crowd-sourcing. Specifically, the annotators were presented with a four-word tuple each time, and asked to select the word with the highest and lowest valence/arousal/dominance, respectively. The best-worst scaling technique was then used to aggregate the annotations: the score for a word is the proportion of times it was chosen as the highest valence/arousal/dominance minus that as the lowest valence/arousal/dominance. The scores for each emotion dimension were then linearly mapped into the interval [0, 1].

With the help of affective norms, we can exploit datasets of different emotion representations for TL, by converting categorical emotion labels into dimensional ones. The affective norms of different languages demonstrate relatively high correlations \cite{warriner2013norms}, suggesting the feasibility for cross-linguistic transfer. Table~\ref{tab:nrc-dict} presents the dimensional emotion scores of some emotion categories in the NRC Lexicon.

\begin{table}[htbp]\centering
\renewcommand{\arraystretch}{1.5}
\caption{Valence, Arousal and Dominance scores of eight typical emotions in the NRC Lexicon.}
\begin{tabular}{cccc} \hline
           & Valence & Arousal & Dominance \\ \hline
Angry      & 0.122   & 0.830   & 0.604     \\
Happy      & 1.000   & 0.735   & 0.772     \\
Sad        & 0.225   & 0.333   & 0.149     \\
Disgusted  & 0.051   & 0.773   & 0.274     \\
Fearful    & 0.083   & 0.482   & 0.278     \\
Surprised  & 0.784   & 0.855   & 0.539     \\
Frustrated & 0.080   & 0.651   & 0.255     \\
Neutral    & 0.469   & 0.184   & 0.357     \\ \hline
\end{tabular} \label{tab:nrc-dict}
\end{table}

\subsection{Cross-Task Inconsistency (CTI)}\label{subsec:cti}

\begin{figure*}[htbp]\centering
\includegraphics[width=0.85\linewidth,clip]{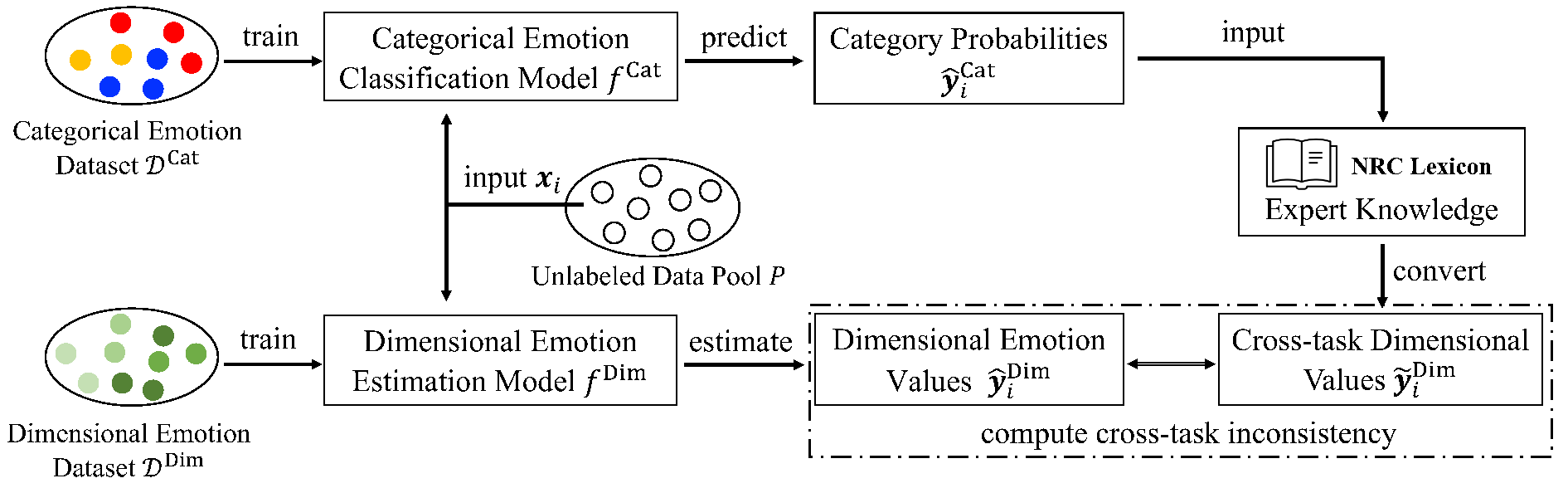}
\caption{Flowchart for computing the CTI.} \label{fig:cti-flowchart}
\end{figure*}

Fig.~\ref{fig:cti-flowchart} shows the flowchart for computing the CTI, an informativeness measure of the prediction inconsistency between two tasks.

First, we construct the emotion classification and estimation models, $f^\text{Cat}$ and $f^\text{Dim}$, using $\mathcal{D}^\text{Cat}$ and $\mathcal{D}^\text{Dim}$ respectively. For an unlabeled sample $\bm{x}_i\in P$, the prediction probabilities for the categorical emotions are:
\begin{align}
	\hat {\bm{y}}_i^\text{Cat}=f^\text{Cat}(\bm{x}_i), 	
	\label{eq:pred_cat_prob}
\end{align}
where each element $\hat{y}_i^e \in \hat {\bm{y}}_i^\text{Cat}$ is the probability of Emotion $e\in E$.

The estimated dimensional emotion values are:
 \begin{align}
	\hat {\bm{y}}_i^\text{Dim}=f^\text{Dim}(\bm{x}_i),
	\label{eq:pred_dim_val}
\end{align}
where $\hat{y}_i^{dim} \in \hat {\bm{y}}_i^\text{Dim}$ is the score of Dimension $dim\in D$.

Utilizing prior knowledge of affective norms, we map the predicted categorical emotion probabilities to dimensional emotion values:
\begin{align}
	\tilde {\bm{y}}_i^\text{Dim}=\sum _{e\in E}{\hat{y}_i^e \cdot \text{NRC}[e]}, 		\label{eq:map_dim_val}
\end{align}
where NRC[$e$] denotes the scores of Emotion $e$ from the NRC Lexicon.

The CTI is then computed as:
\begin{align}
	\mathcal{I}_i=\left\|\hat {\bm{y}}_i^\text{Dim}-\tilde {\bm{y}}_i^\text{Dim}\right\|_2. 	\label{eq:cti}
\end{align}

A high CTI indicates that the two models have a high disagreement, i.e., the corresponding unlabeled sample has high uncertainty (informativeness). Those samples with high CTIs may have inaccurate probability predictions on emotion categories or imprecise estimations of emotion primitives. Labeling these informative samples can expedite the model learning. Besides, the affective norms only represent the most typical (or average) emotion primitives of each category. In practice, a categorical emotion's dimensional primitives may have a large variance. Its corresponding samples are also likely to have high CTIs, and annotating them can increase the sample diversity.

Thus, the sample $\bm{x}_q$ selected for labeling for the target task is:
\begin{align}
	q=\mathop{\arg \max}_{\bm{x_i}\in P} \mathcal{I}_i.
	\label{eq:cti-al}
\end{align}

CTI can be integrated with other useful within-task AL indicators for better performance, as introduced next.

\subsection{CTIAL for Cross-Task Transfer from DEE to CEC}

Uncertainty is a frequently used informativeness metric for AL in classification~\cite{Settles2009}. We consider two typical uncertainty measures in CEC: information entropy and prediction confidence.

The information entropy, also called Shannon entropy \cite{Shannon1948}, is:
\begin{align}
	\text{H}_i=-\sum _{e\in E}\hat{y}_i^e\cdot \log{\hat{y}_i^e}.
	\label{eq:entropy}
\end{align}
A large entropy indicates high uncertainty~\cite{settles2008analysis}.

The prediction confidence directly reflects how certain the classifier is about its prediction:
\begin{align}
	\text{Conf}_i=\max _{e\in E}\hat{y}_i^e.
	\label{eq:confidence}
\end{align}
A low confidence indicates high uncertainty~\cite{culotta2005reducing}.

Considering the uncertainty and CTI simultaneously, we select the sample $\bm{x}_q$ by:
\begin{align}
	q=\mathop{\arg\max}_{\bm{x_i}\in P} \mathcal{I}_i\cdot \text{H}_i,
	\label{eq:cti-entropy}
\end{align}
or by:
\begin{align}
	q=\mathop{\arg\max}_{\bm{x_i}\in P} \frac{\mathcal{I}_i}{\text{Conf}_i},
	\label{eq:cti-confidence}
\end{align}
and query for its class.

The pseudo-code of CTIAL for cross-task transfer from DEE to CEC is given in Algorithm~\ref{alg:CTIAL-CEC}.

\begin{algorithm}[htbp]
 	\KwIn{
 	Source dataset with dimensional emotion values $\mathcal{D}^\text{Dim}=\{(\bm{x}_i, \bm{y}_i^\text{Dim})\}_{i=1}^{N_\text{Dim}}$\;
 	\hspace*{10mm}Target dataset, including data with categorical
 	\hspace*{10mm}emotion labels $\mathcal{D}^\text{Cat}=\{(\bm{x}_i, \bm{y}_i^\text{Cat})\}_{i=1}^{N_\text{Cat}}$ and
  	\hspace*{10mm}unlabeled data pool $P=\{\bm{x}_i\}_{i=1}^{N_P}$\;
  	\hspace*{10mm}$K$, number of samples to be queried\;
  	\hspace*{10mm}Uncertainty measure, entropy or confidence.}
 	\KwOut{Emotion classification model $f^\text{Cat}$.}
 	Train the model $f^\text{Cat}$ on $\mathcal{D}^\text{Cat}$ and $f^\text{Dim}$ on $\mathcal{D}^\text{Dim}$\;
 	Estimate dimensional emotion values $\{\hat{\bm{y}}_i^\text{Dim}\}_{i=1}^{N_P}$ of $P$ using \eqref{eq:pred_dim_val}\;
 	
 	\For{$k=1:K$}{
	
 	Estimate emotion category probabilities $\{\hat{\bm{y}}_i^\text{Cat}\}_{i=1}^{N_P}$ of $P$ using \eqref{eq:pred_cat_prob}\;
  	Map $\{\hat{\bm{y}}_i^\text{Cat}\}_{i=1}^{N_P}$ into the dimensional emotion space and obtain $\{\tilde {\bm{y}}_i^\text{Dim}\}_{i=1}^{N_P}$ using \eqref{eq:map_dim_val}\;
  	Calculate CTI $\{\mathcal{I}_i\}_{i=1}^{N_P}$ using \eqref{eq:cti}\;
  	\uIf{the uncertainty measure is entropy}
  	{Compute entropy $\{\text{H}_i\}_{i=1}^{N_P}$ using \eqref{eq:entropy}\;
  	Select sample $\bm{x}_q$ using \eqref{eq:cti-entropy}\;
  	}
  	\Else{
  	\tcp*[h]{the uncertainty measure is confidence}
  	
  	Compute confidence $\{\text{Conf}_i\}_{i=1}^{N_P}$ using \eqref{eq:confidence}\;
  	Select sample $\bm{x}_q$ using \eqref{eq:cti-confidence}\;}

 	Query for the emotion category $\bm{y}_q^\text{Cat}$ of $\bm{x}_q$\;
 	$\mathcal{D}^\text{Cat}\leftarrow \mathcal{D}^\text{Cat}\cup (\bm{x}_q, \bm{y}_q^\text{Cat})$\;
 	$P\leftarrow P\backslash \bm{x}_q$; \quad $N_P\leftarrow N_P-1$\;
	Update $f^\text{Cat}$ on $\mathcal{D}^\text{Cat}$\;
 	}
 \caption{CTIAL for cross-task transfer from DEE to CEC.}  	
 \label{alg:CTIAL-CEC}
 \end{algorithm}

\subsection{CTIAL for Cross-Task Transfer from CEC to DEE}

Wu \emph{et al.} \cite{wu2019active,wu2019affect} employed greedy sampling in both feature and label spaces for sample selection and verified the importance of diversity in AL for regression. Their multi-task improved greedy sampling (MTiGS) \cite{wu2019affect} computes the distance between an unlabeled sample $\bm{x}_i$ and the labeled sample set $\mathcal{D}^\text{Dim}$ as:
 \begin{align}
	d_i=\min _{\bm{x}_j\in \mathcal{D}^\text{Dim}}\|\bm{x}_i-\bm{x}_j\|_2\cdot \prod_{dim\in D} |\hat{y}_i^{dim}-y_j^{dim}|.
	\label{eq:igs}
\end{align}
To weight the feature and label spaces equally, we slightly modify MTiGS to:
\begin{align}
	d_i^{\prime}=\min _{\bm{x}_j\in \mathcal{D}^\text{Dim}}\|\bm{x}_i-\bm{x}_j\|_2\cdot \|\hat{\bm{y}}_i^\text{Dim}-\bm{y}_j^\text{Dim}\|_2.
	\label{eq:igs-our}
\end{align}

Considering CTI (informativeness) and MTiGS (diversity) together, we select the sample $\bm{x}_q$ by:
\begin{align}
	q=\mathop{\arg\max}_{\bm{x_i}\in P} \mathcal{I}_i\cdot d_i^{\prime},
	\label{eq:cti-igs}
\end{align}
and query for its dimensional emotion values.

The pseudo-code of CTIAL for DEE is shown in Algorithm~\ref{alg:CTIAL-DEE}.

\begin{algorithm}[htbp]
 	\KwIn{
 	Source dataset with categorical emotion labels $\mathcal{D}^\text{Cat}=\{(\bm{x}_i, \bm{y}_i^\text{Cat})\}_{i=1}^{N_\text{Cat}}$\;
 	\hspace*{10mm}Target dataset, including data with dimensional
 	\hspace*{10mm}emotion values $\mathcal{D}^\text{Dim}=\{(\bm{x}_i, \bm{y}_i^\text{Dim})\}_{i=1}^{N_\text{Dim}}$ and
 	\hspace*{10mm}unlabeled data pool $P=\{\bm{x}_i\}_{i=1}^{N_P}$\;
  	\hspace*{10mm}$K$, number of samples to be queried.}
 	\KwOut{Emotion estimation model $f^\text{Dim}$.}

 	Train the model $f^\text{Cat}$ on $\mathcal{D}^\text{Cat}$ and $f^\text{Dim}$ on $\mathcal{D}^\text{Dim}$\;
 	Estimate category probabilities $\{\hat{\bm{y}}_i^\text{Cat}\}_{i=1}^{N_P}$ of $P$ by \eqref{eq:pred_cat_prob}\;
 	Map $\{\hat{\bm{y}}_i^\text{Cat}\}_{i=1}^{N_P}$ into the dimensional emotion space and obtain $\{\tilde {\bm{y}}_i^\text{Dim}\}_{i=1}^{N_P}$ using \eqref{eq:map_dim_val}\;
 	
 	\For{$k=1:K$}{
 	Estimate dimensional emotion values $\{\hat{\bm{y}}_i^\text{Dim}\}_{i=1}^{N_P}$ of $P$ using \eqref{eq:pred_dim_val}\;
  	Calculate CTI $\{\mathcal{I}_i\}_{i=1}^{N_P}$ using \eqref{eq:cti}\;
 	Compute the distance $\{d_i^{\prime}\}_{i=1}^{N_P}$ between samples in $P$ and $\mathcal{D}^\text{Dim}$ using \eqref{eq:igs-our}\;
 	Select sample $\bm{x}_q$ using \eqref{eq:cti-igs}\;
 	Query for the dimensional emotion values $\bm{y}_q^\text{Dim}$ of $\bm{x}_q$\;
 	$\mathcal{D}^\text{Dim}\leftarrow \mathcal{D}^\text{Dim}\cup (\bm{x}_q, \bm{y}_q^\text{Dim})$\;
 	$P\leftarrow P\backslash \bm{x}_q$; \quad $N_P\leftarrow N_P-1$\;
 	Update $f^\text{Dim}$ on $\mathcal{D}^\text{Dim}$\;
 	}
 	\caption{CTIAL for cross-task transfer from CEC to DEE.}  	
 	\label{alg:CTIAL-DEE}

 \end{algorithm}

\subsection{Domain Adaptation in Cross-Corpus Transfer}

CTIAL assumes the source model can make reliable predictions for the target dataset. However, speech emotion recognition corpora vary according to whether they are acted or spontaneous, collected in lab or in the wild, and so on. These discrepancies may cause a model trained on one corpus to perform poorly on another, violating the underlying assumption in CTIAL.

To enable cross-corpus transfer, we adopt two classical domain adaptation approaches, transfer component analysis (TCA)~\cite{pan2010domain} and balanced distribution adaptation~(BDA)~\cite{wang2017balanced}.

When the source task is CEC, BDA jointly aligns both the marginal distributions and the class-conditional distributions with a balance factor that adjusts the weights of the two corresponding terms in the objective function. Specifically, the source model assigns pseudo-labels for the unlabeled target dataset. The source and target datasets' average features and their corresponding class' average features are aligned. The adapted features of the source dataset are then used to update the source model. This process iterates till convergence.

When the source task is DEE, TCA adapts the marginal distributions of the source and target datasets by reducing the distance between their average features. BDA is not used here, since the class-conditional probabilities are difficult to compute in regression problems.

After domain adaptation, the model trained on the source dataset is applied to $P$ to obtain the predictions for the source task. More details on the implementations will be introduced in Section~\ref{subsec:datasets}.

\section{Experimental Setup} \label{sec:setup}

This section describes the datasets and the experimental setup.

\subsection{Datasets and Feature Extraction}
\label{subsec:datasets}

Three public speech emotion datasets, IEMOCAP (Interactive Emotional Dyadic Motion Capture Database) \cite{busso2008iemocap} of semi-authentic emotions, MELD~(Multimodal EmotionLines Dataset) \cite{poria2019meld} of semi-authentic emotions, and VAM (Vera am Mittag; Vera at Noon in English) \cite{grimm2008vera} of authentic emotions, were used to verify the proposed CTIAL. Specifically, we performed experiments of within-corpus transfer on IEMOCAP, and cross-corpus transfer from VAM and MELD to IEMOCAP.

IEMOCAP is a multi-modal dataset annotated with both categorical and dimensional emotion labels. Audio data from five emotion categories (angry, happy\footnote{The `happy' class contains data of `happy' and `excited' in the original annotation as in multiple previous works.}, sad, frustrated and neutral with 289, 947, 608, 971 and 1099 samples, respectively) in spontaneous sessions were used in the within-corpus experiments and cross-corpus transfer from DEE on VAM. In cross-corpus transfer from CEC on MELD, only four overlapping classes (angry, happy, sad and neutral) were used. The valence, arousal and dominance annotations are in [1, 5], so we linearly rescaled the scores in the NRC Lexicon from [0, 1] to [1, 5].

MELD is a multi-modal dataset collected from the TV series \emph{Friends}. Each utterance was labeled by five annotators from seven classes (anger, disgust, sadness, joy, surprise, fear, and neutral), and majority vote was applied to generate the ground-truth labels. We only used the four overlapping categories with IEMOCAP in the training set (angry, happy, sad and neutral with 1109, 1743, 683 and 4709 samples, respectively).

VAM consists of 947 utterances collected from 47 guests (11m/36f) in a German TV talk-show \emph{Vera am Mittag}. Each sentence was annotated by 6 or 17 evaluators for valence, arousal and dominance values, and the weighted average was used to obtain the ground-truth labels. We linearly rescaled their values from [-1, 1] to [1, 5] to match the range of IEMOCAP.

The wav2vec~2.0 model \cite{baevski2020wav2vec}, pre-trained on 960 hours of unlabeled audio from the LibriSpeech dataset \cite{panayotov2015librispeech} and fine-tuned for automatic speech recognition on the same audio with transcripts, was used for feature extraction. For each audio segment, we took the average output of 12 transformer encoder layers, and averaged these features again along the time axis, obtaining a 768-dimensional feature for each utterance.

In cross-task AL, we combined the source and target datasets and used principal component analysis to reduce the feature dimensionality (maintaining 90\% variance). In cross-corpus transfer, an additional feature adaptation step was performed using TCA or BDA to obtain more accurate source task predictions for samples in $P$.

Before applying AL to $P$, we applied principal component analysis to the original 768-dimensional features of the target dataset in both cross-task AL and within-task AL baselines.

\subsection{Experimental Setup}
We used logistic regression (LR) and ridge regression (RR) as the base model for CEC and DEE, respectively. The weight of the regularization term in each model, i.e., $1/C$ in LR and $\alpha$ in RR, was chosen from $\{1, 5, 10, 50, 1e2, 5e2, 1e3, 5e3\}$ by three-fold cross-validation on the corresponding training data.

In cross-corpus transfer experiments, feature dimensionality in TCA and BDA was set to 30 and 40, respectively. BDA used 10 iterations to estimate the class labels of the target dataset, align the marginal and conditional distributions, and update the classifier. The balance factor was selected from $\{0.1, 0.2, ..., 0.9\}$ to minimize the sum of the maximum mean discrepancy metrics~\cite{borgwardt2006integrating} on all data and in each class.

\subsection{Performance Evaluation}

We aimed to classify all samples in the target dataset, some by human annotators and the remaining by the trained classifier. Therefore, we evaluated the classification performance on the entire target dataset, including the manually labeled samples and $P$. Specifically, we concatenated the ground-truth labels $\{\bm{y}_i^\text{Cat}\}_{i=1}^{N_\text{Cat}}$ of $\mathcal{D}^\text{Cat}$ and the predictions $\{\hat{\bm{y}}_i^\text{Cat}\}_{i=1}^{N_P}$ of $P$ to compute the performance metrics.

In within-corpus transfer, either from DEE to CEC or from CEC to DEE on IEMOCAP, each time we used two sessions as the source dataset and the rest three sessions as the target dataset, resulting in 10 source-target dataset partitions. Experiments on each dataset partition were repeated three times with different initial labeled samples.

Cross-corpus experiments considered transfer from the source DEE task on VAM to the target CEC task on IEMOCAP and from the source CEC task on MELD to the target DEE task on IEMOCAP. Experiments were repeated 10 times with different initial labeled sample sets.

In each run of the experiment, the initial 20 labeled samples were randomly selected. In the subsequent 200 iterations of sample selection, one sample was chosen for annotation at a time.

Balanced classification accuracy (BCA), i.e., the average per-class accuracies, was used as the performance measure in CEC since IEMOCAP has significant class-imbalance. Root mean squared error (RMSE) and correlation coefficient (CC) were used as performance measures in DEE.

\section{Experiments on Cross-Task Transfer from DEE to CEC} \label{sec:exp_CEC}

This section presents the results in cross-task transfer from DEE to CEC. The DEE and CEC tasks could be from the same corpus, or different ones.

\subsection{Algorithms}
We compared the performance of the following sample selection strategies in cross-task transfer from DEE to CEC:
\begin{enumerate}
	\item Random sampling (\texttt{Rand}), which randomly selects $K$ samples to annotate.
	\item Entropy (\texttt{Ent})~\cite{settles2008analysis}, an uncertainty-based within-task AL approach that selects samples with maximum entropy computed by \eqref{eq:entropy}.
	\item Least confidence (\texttt{LC})~\cite{culotta2005reducing}, an uncertainty-based within-task AL approach that selects samples with minimum prediction confidence computed by \eqref{eq:confidence}.
	\item Multi-task iGS on the source task (\texttt{Source MTiGS}), which performs AL according to the informativeness metric in the source DEE task. Specifically, we compute the distance between samples $\bm{x}_i\in P$ and $\bm{x}_j\in \mathcal{D}^\text{Cat}$ by~\eqref{eq:igs} and select samples with the maximum distance. Without any dimensional emotion labels of $\mathcal{D}^\text{Cat}$, we replace the true label $y_j^{dim}$ with $\hat{y}_j^{dim}$ estimated from the source model. This simple cross-task AL baseline directly uses the knowledge from the source task.
	\item \texttt{CTIAL}, which selects samples with the maximum CTI by~\eqref{eq:cti-al}.
	\item \texttt{Ent-CTIAL}, which integrates entropy with CTI as introduced in Algorithm~\ref{alg:CTIAL-CEC}.
	\item \texttt{LC-CTIAL}, which integrates the prediction confidence with CTI as introduced in Algorithm~\ref{alg:CTIAL-CEC}.
\end{enumerate}

\subsection{Effectiveness of CTIAL}

\begin{figure}[htbp]\centering
\subfigure[]{\includegraphics[width=0.9\linewidth,clip] {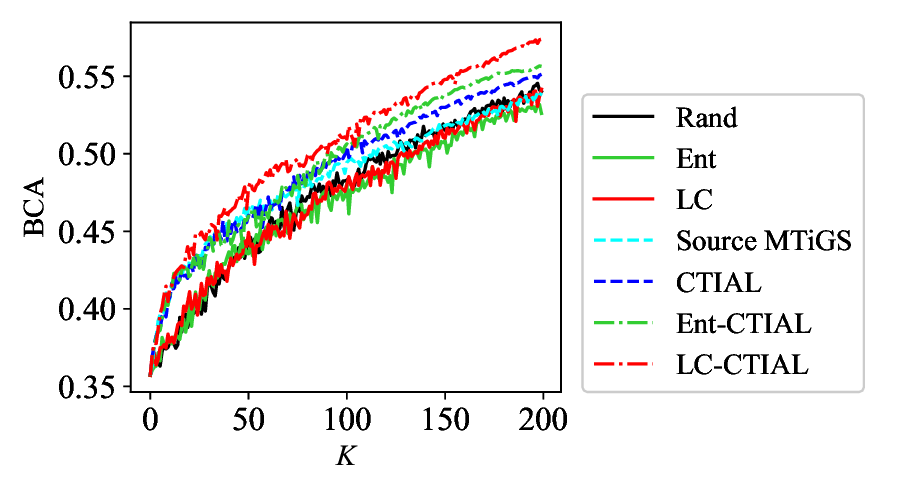}}
\subfigure[]{\includegraphics[width=0.9\linewidth,clip] {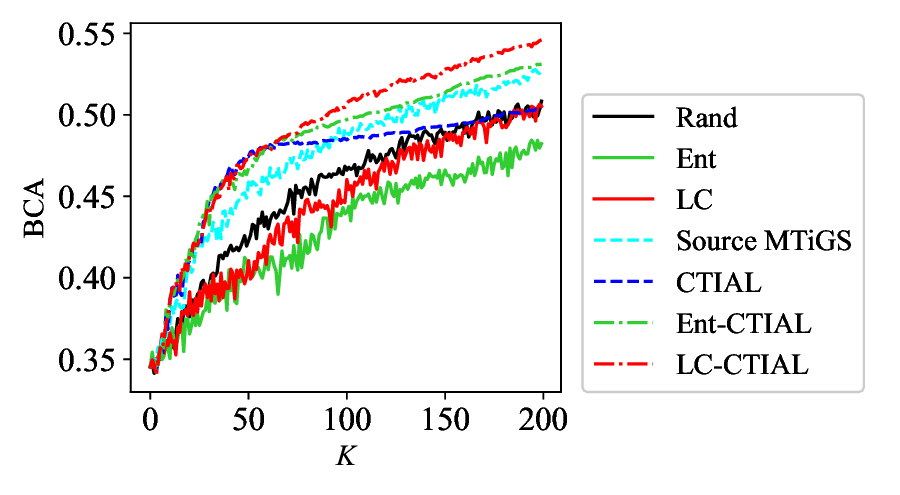}}
\caption{Average BCAs of different sample selection approaches in cross-task transfer from DEE to CEC. (a) Within-corpus transfer from DEE on IEMOCAP to CEC on IEMOCAP; and, (b) cross-corpus transfer from DEE on VAM to CEC on IEMOCAP. $K$ is the number of samples to be queried in addition to the initial labeled ones.} \label{fig:iemocap-cec}
\end{figure}

\begin{figure}[htbp]\centering
\subfigure[]{\includegraphics[width=0.65\linewidth,clip] {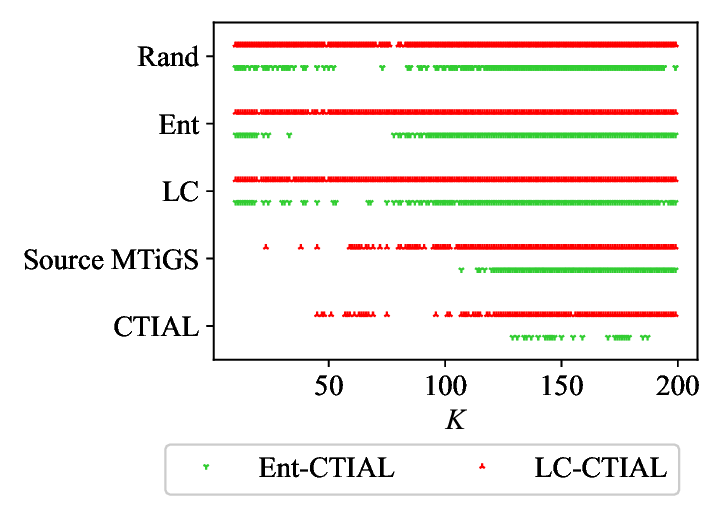}}
\subfigure[]{\includegraphics[width=0.65\linewidth,clip] {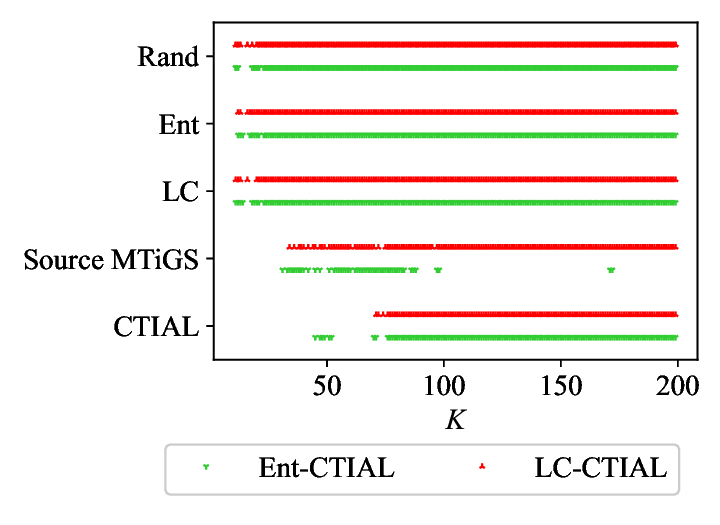}}
\caption{
Statistical significance of the performance improvements of Ent-CTIAL and LC-CTIAL over the other approaches. (a) Within-corpus transfer from DEE on IEMOCAP to CEC on IEMOCAP; and, (b) cross-corpus transfer from DEE on VAM to CEC on IEMOCAP. $K$ is the number of samples to be queried in addition to the initial labeled ones. The vertical axis denotes the approaches in comparison with Ent-CTIAL or LC-CTIAL in Wilcoxon signed-rank tests. The red and green markers were placed at where the adjusted $p$-values were smaller than 0.05.} \label{fig:wilcoxon-cec}
\end{figure}

Fig.~\ref{fig:iemocap-cec} shows the average BCAs in within- and cross-corpus cross-task transfers from DEE to CEC. To examine if the performance improvements of the integration of the uncertainty-based approaches and CTIAL were statistically significant, Wilcoxon signed-rank tests with Holm's $p$-value adjustment~\cite{holm1979simple} were performed on the results in each AL iteration. The test results are shown in Fig.~\ref{fig:wilcoxon-cec}.

Figs.~\ref{fig:iemocap-cec} and~\ref{fig:wilcoxon-cec} demonstrate that:
\begin{enumerate}
	\item As the number of labeled samples increased, classifiers in all approaches became more accurate, achieving higher BCAs.
	\item Between the two uncertainty-based AL approaches, \texttt{LC} outperformed \texttt{Ent}. However, both may not outperform \texttt{Rand}, suggesting that only considering the uncertainty may not be enough.
	\item \texttt{Source MTiGS} that performed AL according to the source DEE task achieved higher BCAs than \texttt{Rand} and the two within-task AL approaches, because MTiGS increased the feature diversity, which was also useful for the target CEC task. Additionally, MTiGS increased the diversity of the dimensional emotion labels, which in turn increased the diversity of the categorical emotion labels.
	\item \texttt{CTIAL} outperformed \texttt{Rand}, the two within-task AL approaches (\texttt{Ent} and \texttt{LC}), and the cross-task AL baseline \texttt{Source MTiGS} when $K$ was small, demonstrating that the proposed CTI measure properly exploited the relationship between the CEC and DEE tasks and effectively utilized knowledge from the source task. However, in cross-corpus transfer, \texttt{CTIAL} and \texttt{Rand} had similar performance when $K$ was large. The reason may be that domain shift limited the performance of the source DEE models on the target dataset, which further resulted in inaccurate CTI calculation and degraded AL performance.
	\item Integrating CTI with uncertainty can further enhance the classification accuracies by simultaneously considering within- and cross-task informativeness: both \texttt{LC-CTIAL} and \texttt{Ent-CTIAL} statistically significantly outperformed the other baselines.
	\item \texttt{LC-CTIAL} generally achieved the best performance among all seven approaches.
\end{enumerate}

\subsection{Effectiveness of TCA}
CTIAL assumes that the model trained on the source dataset is reliable for the target dataset. We performed within-corpus and cross-corpus experiments to verify this assumption.

\begin{figure}[htbp]\centering
	\includegraphics[width=\linewidth, clip]{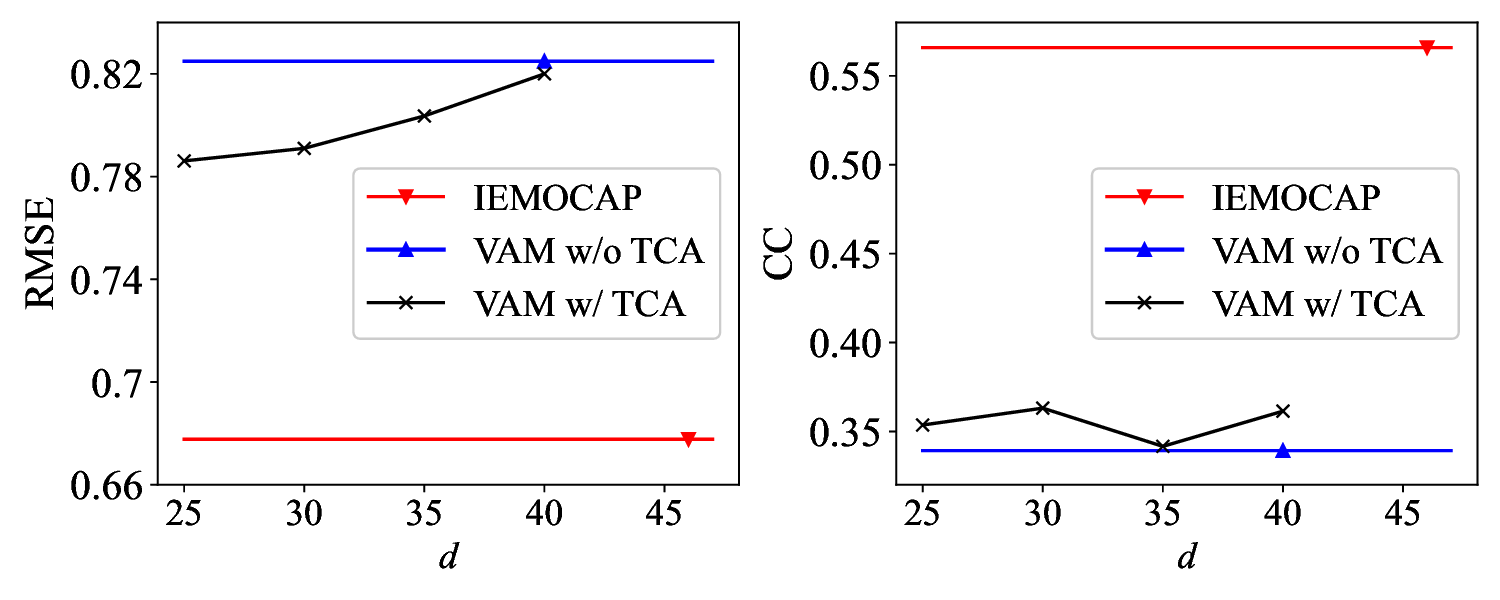}
	\caption{Average RMSEs and CCs in valence, arousal and dominance estimation on IEMOCAP in within-corpus transfer (DEE on IEMOCAP to DEE on IEMOCAP; red line), direct cross-corpus transfer (DEE on VAM to DEE on IEMOCAP; blue line), and cross-corpus transfer using TCA (DEE on VAM to DEE on IEMOCAP; black curve). $d$ is the feature dimensionality. The markers on the red and blue lines mean that the feature dimensionality after principal component analysis was 46 and 40, respectively.} \label{fig:tca-dee}
\end{figure}

Fig.~\ref{fig:tca-dee} shows the average RMSEs and CCs in valence, arousal and dominance estimation on IEMOCAP in within- and cross-corpus transfers. In within-corpus transfer, the average RMSE and CC were 0.6667 and 0.5659, respectively. Although models trained on VAM were inferior to those on IEMOCAP, TCA still achieved lower RMSEs and higher CCs than direct transfer.

Generally, the assumption that the source model is reliable was satisfied in both within-corpus transfer and more challenging cross-corpus transfer, with the help of TCA.

\section{Experiments on Cross-Task Transfer from CEC to DEE} \label{sec:exp_DEE}

This section presents the experiment results in cross-task transfer from CEC to DEE, where the two tasks may be from the same or different datasets.

\subsection{Algorithms}

The following approaches were compared in transferring from CEC to DEE:
\begin{enumerate}
	\item Direct mapping by NRC Lexicon~(\texttt{NRC Mapping}), where the dimensional emotion estimates of all the samples are obtained by~\eqref{eq:map_dim_val}. This non-AL approach only utilizes the information of the source CEC task and domain knowledge.
	\item Random sampling (\texttt{Rand}), which randomly selects $K$ samples to annotate.
	\item Multi-task iGS (\texttt{MTiGS})~\cite{wu2019affect}, a diversity-based within-task AL approach that selects samples with the furthest distance to labeled data computed by~\eqref{eq:igs}.
	\item Least confidence on the source task (\texttt{Source LC}), which selects samples with the minimum source model prediction confidence defined by \eqref{eq:confidence}. This cross-task AL baseline depends solely on the source CEC model.
	\item Cross-task iGS~(\texttt{CTiGS}), a variant of \texttt{MTiGS} that further considers the information in the source CEC task. Specifically, we first obtain the predicted emotion categories for the target data using the source model. The distance calculation is only conducted between unlabeled and labeled samples predicted with the same emotion category. For an emotion category with only unlabeled samples, we calculate the distance between them and all the labeled samples using~\eqref{eq:igs}, as in \texttt{MTiGS}. The subsequent sample selection process is the same as \texttt{MTiGS}.
	\item \texttt{CTIAL}, which selects samples with the maximum CTI by~\eqref{eq:cti-al}.
	\item \texttt{MTiGS-CTIAL}, which integrates MTiGS and CTI as introduced in Algorithm~\ref{alg:CTIAL-DEE}.
\end{enumerate}

\subsection{Effectiveness of CTIAL}

\begin{figure*}[htbp]\centering
\subfigure[]{\includegraphics[width=0.95\linewidth,clip]{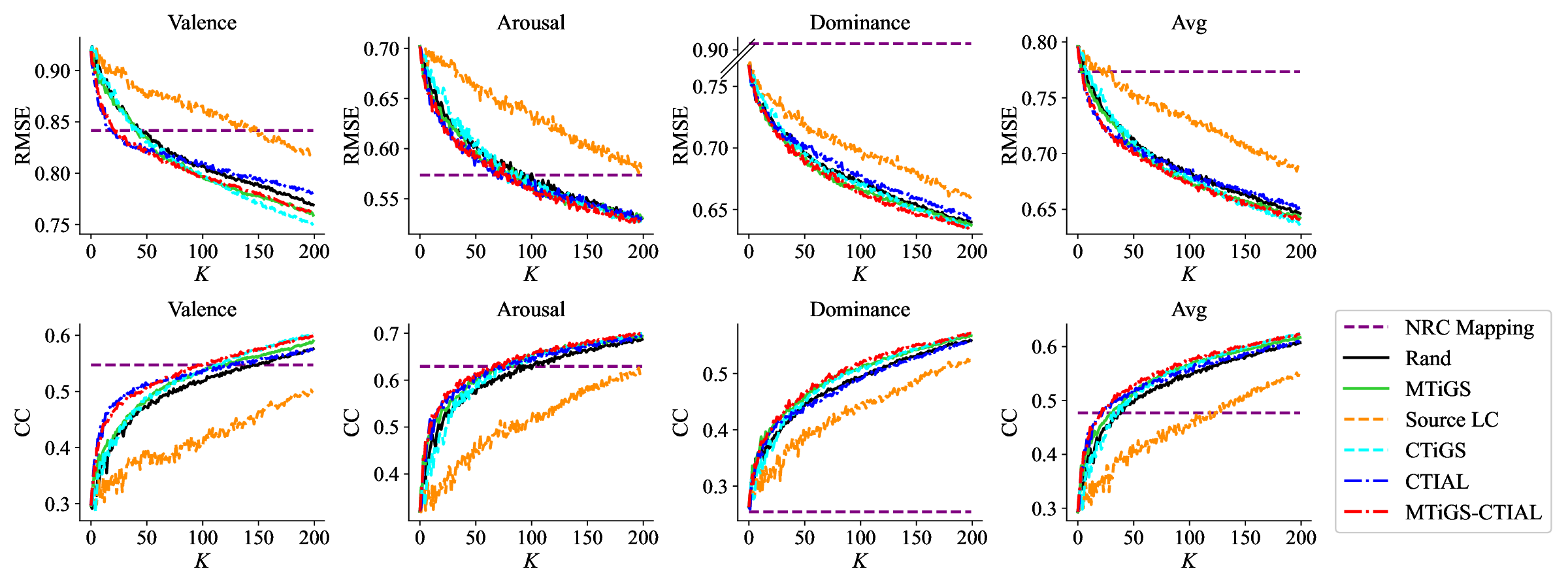}}
\subfigure[]{\includegraphics[width=0.95\linewidth,clip]{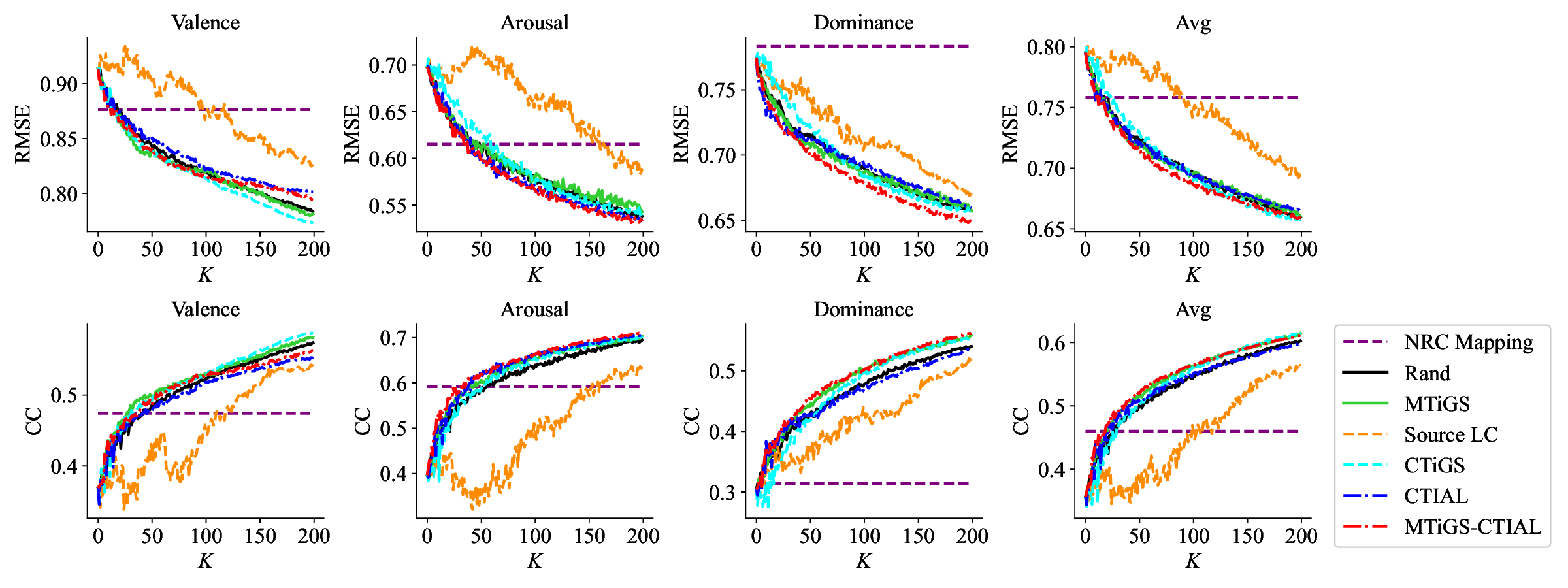}}
\caption{Average RMSEs and CCs of different sample selection approaches in cross-task transfer from CEC to DEE. (a) Within-corpus transfer from CEC on IEMOCAP to DEE on IEMOCAP; and, (b) cross-corpus transfer from CEC on MELD to DEE on IEMOCAP. $K$ is the number of samples to be queried in addition to the initial labeled ones.} \label{fig:iemocap-dee}
\end{figure*}

\begin{figure}[htbp]\centering
\subfigure[]{\includegraphics[width=0.6\linewidth,clip] {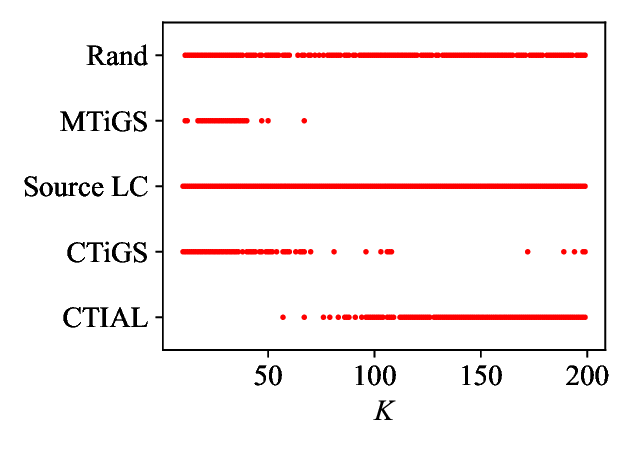}}
\subfigure[]{\includegraphics[width=0.6\linewidth,clip] {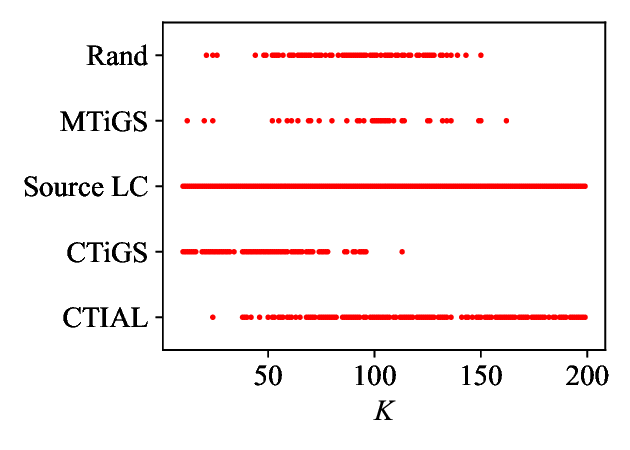}}
\caption{Statistical significance of the performance improvements of MTiGS-CTIAL over the other approaches. (a) Within-corpus transfer from CEC on IEMOCAP to DEE on IEMOCAP; and, (b) cross-corpus transfer from CEC on MELD to DEE on IEMOCAP. $K$ is the number of samples to be queried in addition to the initial labeled ones. The vertical axis denotes the approaches in comparison with MTiGS-CTIAL in Wilcoxon signed-rank tests. The red markers were placed at where the adjusted $p$-values were smaller than 0.05.} \label{fig:wilcoxon-dee}
\end{figure}

Fig.~\ref{fig:iemocap-dee} shows the average performance in DEE on IEMOCAP using different sample selection approaches. Wilcoxon signed-rank tests with Holm's $p$-value adjustment~\cite{holm1979simple} were again used to examine if the performance improvements of MTiGS-CTIAL over other approaches were statistically significant in each AL iteration. The test results are shown in Fig.~\ref{fig:wilcoxon-dee}.

Figs.~\ref{fig:iemocap-dee} and \ref{fig:wilcoxon-dee} shows that:
\begin{enumerate}
	\item The regression models in all approaches performed better as the number of labeled samples increased, except the non-AL baseline \texttt{NRC Mapping}.
	\item \texttt{NRC Mapping} had poor performance, especially in dominance estimation. Models trained on only a small amount of labeled data outperformed \texttt{NRC Mapping}, regardless of the sample selection approach. As stated in Section~\ref{subsec:cti}, each emotion category was mapped to a single tuple of dimensional emotion values according to the affective norms. In practice, samples belonging to the same emotion category may have diverse emotion primitives. Direct mapping oversimplified the relationship between categorical emotions and dimensional emotions.
	\item \texttt{MTiGS} on average achieved lower RMSEs and higher CCs than \texttt{Rand}. However, it performed similarly to \texttt{Rand} for some emotion primitives.
	\item \texttt{Source LC} performed much worse than \texttt{Rand} because it only considered the classification information in the source task but ignored the characteristics of the target task. Samples with low classification confidence may not be very useful to the target regression tasks.
	\item \texttt{CTiGS} initially performed worse than \texttt{MTiGS}, but gradually outperformed on some dimensions as the number of labeled samples increased. The reason is that \texttt{CTiGS} emphasizes the within-class sample diversity, whereas \texttt{MTiGS} focuses on the global sample diversity. The latter is more beneficial for the regression models to learn global patterns, which is important when labeled samples are rare. As $K$ increases, enriching the within-class sample diversity facilitates the global models to learn local patterns.
	\item In within-corpus transfer, \texttt{CTIAL} outperformed both \texttt{Rand} and \texttt{MTiGS} initially, but gradually became inferior to \texttt{Rand}. In cross-corpus transfer, \texttt{CTIAL} performed better than \texttt{Rand} on arousal but worse on valence and dominance. The possible reason is similar to that in transfer from DEE to CEC: the source classification models were not accurate enough, resulting in inaccurate CTI calculation and unsatisfactory AL performance.
	\item \texttt{MTiGS-CTIAL} achieved the best overall performance in both within- and cross-corpus transfers, indicating that considering both CTI and diversity in AL helped improve the regression performance even when the CTI was not very accurate.
\end{enumerate}

\subsection{Effectiveness of BDA}

\begin{figure}[htbp]\centering
	\includegraphics[width=0.6\linewidth, clip]{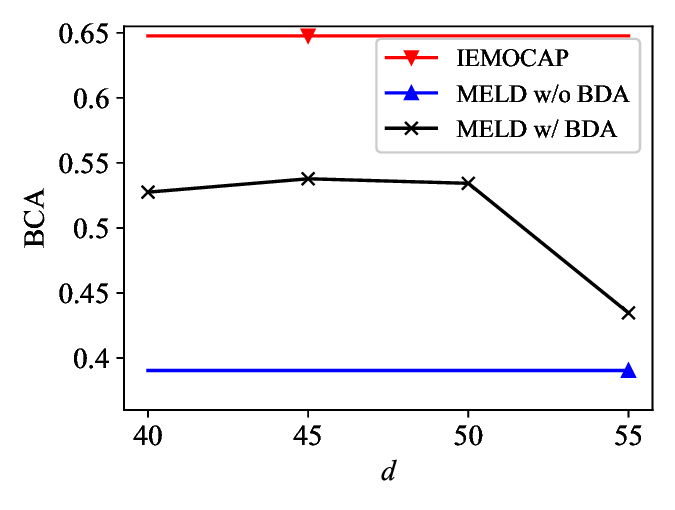}
	\caption{BCAs in CEC on IEMOCAP in within-corpus transfer (CEC on IEMOCAP to CEC on IEMOCAP; red line), direct cross-corpus transfer (CEC on MELD to CEC on IEMOCAP; blue line), and cross-corpus transfer using BDA (CEC on MELD to CEC on IEMOCAP; black curve). $d$ denotes the feature dimensionality. The markers on the red and blue lines mean that the feature dimensionality after principal component analysis was 45 and 55, respectively.}
\label{fig:bda-cec}
\end{figure}

Fig.~\ref{fig:bda-cec} shows the classification results in CEC on IEMOCAP using different training data and projection dimensionality for BDA in cross-corpus transfer. In within-corpus validation, we used two sessions as the training data and the rest three sessions as the test data. The average BCA in 10 training-test partitions for four-class ($E$=\{angry, happy, sad, neutral\}) emotion recognition reached 0.6477. In cross-corpus transfer, directly transferring the model trained on MELD to IEMOCAP only achieved a BCA of 0.3904. BDA boosted the BCA to 0.5378 when setting the projected feature dimensionality to 45.

\section{Conclusions} \label{sec:conclusion}

Human emotions can be described by both categorical and dimensional representations. Usually, training accurate emotion classification and estimation models requires large labeled datasets. However, manual labeling of affective samples is expensive, due to the subtleness and complexity of emotions. This paper integrates AL and TL to reduce the labeling effort. We proposed CTI to measure the prediction inconsistency between the CEC and DEE tasks. To further consider other useful AL metrics, CTI was integrated with uncertainty in CEC and diversity in DEE for enhanced reliability. Experiments on three speech emotion datasets demonstrated the effectiveness of CTIAL in within- and cross-corpus transfers between DEE and CEC tasks. To our knowledge, this is the first work that utilizes prior knowledge on affective norms and data in a different task to facilitate AL for a new task, regardless of whether the two tasks are from the same dataset or not.


\begin{thebibliography}{10}
\providecommand{\url}[1]{#1}
\csname url@samestyle\endcsname
\providecommand{\newblock}{\relax}
\providecommand{\bibinfo}[2]{#2}
\providecommand{\BIBentrySTDinterwordspacing}{\spaceskip=0pt\relax}
\providecommand{\BIBentryALTinterwordstretchfactor}{4}
\providecommand{\BIBentryALTinterwordspacing}{\spaceskip=\fontdimen2\font plus
\BIBentryALTinterwordstretchfactor\fontdimen3\font minus
  \fontdimen4\font\relax}
\providecommand{\BIBforeignlanguage}[2]{{%
\expandafter\ifx\csname l@#1\endcsname\relax
\typeout{** WARNING: IEEEtran.bst: No hyphenation pattern has been}%
\typeout{** loaded for the language `#1'. Using the pattern for}%
\typeout{** the default language instead.}%
\else
\language=\csname l@#1\endcsname
\fi
#2}}
\providecommand{\BIBdecl}{\relax}
\BIBdecl

\bibitem{pantic2000}
M.~Pantic and L.~Rothkrantz, ``Automatic analysis of facial expressions: {T}he
  state of the art,'' \emph{{IEEE} Trans. on Pattern Analysis and Machine
  Intelligence}, vol.~22, no.~12, pp. 1424--1445, 2000.

\bibitem{Kaliouby2004}
R.~E. Kaliouby and P.~Robinson, ``Real-time inference of complex mental states
  from facial expressions and head gestures,'' in \emph{Proc. Int'l Conf. on
  Computer Vision and Pattern Recognition}, Washington {DC}, June 2004, p. 154.

\bibitem{wu2019affect}
D.~Wu and J.~Huang, ``Affect estimation in {3D} space using multi-task active
  learning for regression,'' \emph{IEEE Trans. on Affective Computing},
  vol.~13, no.~41, pp. 16--27, 2022.

\bibitem{drwuPIEEE2023}
D.~Wu, B.-L. Lu, B.~Hu, and Z.~Zeng, ``Affective brain-computer interfaces
  ({aBCIs}): A tutorial,'' \emph{Proc. of the IEEE}, 2023, in press.

\bibitem{Jiang2020}
Y.~Jiang, W.~Li, M.~S. Hossain, M.~Chen, A.~Alelaiwi, and M.~Al-Hammadi, ``A
  snapshot research and implementation of multimodal information fusion for
  data-driven emotion recognition,'' \emph{Information Fusion}, vol.~53, pp.
  209--221, 2020.

\bibitem{Ayata2018}
D.~Ayata, Y.~Yaslan, and M.~E. Kamasak, ``Emotion based music recommendation
  system using wearable physiological sensors,'' \emph{IEEE Trans. on Consumer
  Electronics}, vol.~64, no.~2, pp. 196--203, 2018.

\bibitem{ekman1987universals}
P.~Ekman, W.~V. Friesen, M.~O'sullivan, A.~Chan, I.~Diacoyanni-Tarlatzis,
  K.~Heider, R.~Krause, W.~A. LeCompte, T.~Pitcairn, P.~E. Ricci-Bitti
  \emph{et~al.}, ``Universals and cultural differences in the judgments of
  facial expressions of emotion.'' \emph{Journal of Personality and Social
  Psychology}, vol.~53, no.~4, p. 712, 1987.

\bibitem{mehrabian1980basic}
A.~Mehrabian, \emph{Basic dimensions for a general psychological theory:
  Implications for personality, social, environmental, and developmental
  studies}.\hskip 1em plus 0.5em minus 0.4em\relax Cambridge, MA:
  Oelgeschlager, Gunn \& Hain, 1980.

\bibitem{russell1980circumplex}
J.~A. Russell, ``A circumplex model of affect.'' \emph{Journal of Personality
  and Social Psychology}, vol.~39, no.~6, p. 1161, 1980.

\bibitem{Settles2009}
B.~Settles, ``Active learning literature survey,'' University of
  Wisconsin--Madison, Computer Sciences Technical Report 1648, 2009.

\bibitem{pan2010survey}
S.~J. Pan and Q.~Yang, ``A survey on transfer learning,'' \emph{{IEEE} Trans.
  on Knowledge and Data Engineering}, vol.~22, no.~10, pp. 1345--1359, 2010.

\bibitem{muhammad2017user}
G.~Muhammad and M.~F. Alhamid, ``User emotion recognition from a larger pool of
  social network data using active learning,'' \emph{Multimedia Tools and
  Applications}, vol.~76, no.~8, pp. 10\,881--10\,892, 2017.

\bibitem{zhang2015dynamic}
Y.~Zhang, E.~Coutinho, Z.~Zhang, C.~Quan, and B.~Schuller, ``Dynamic active
  learning based on agreement and applied to emotion recognition in spoken
  interactions,'' in \emph{Proc. of the ACM on Int'l Conf. on Multimodal
  Interaction}, Seattle, Washington, WA, Nov. 2015, pp. 275--278.

\bibitem{wu2019active}
D.~Wu, C.-T. Lin, and J.~Huang, ``Active learning for regression using greedy
  sampling,'' \emph{Information Sciences}, vol. 474, pp. 90--105, 2019.

\bibitem{abdelwahab2019active}
M.~Abdelwahab and C.~Busso, ``Active learning for speech emotion recognition
  using deep neural network,'' in \emph{Proc. Int'l Conf. on Affective
  Computing and Intelligent Interaction}, Cambridge, UK, Sep. 2019, pp. 1--7.

\bibitem{wu2020transfer}
D.~Wu, Y.~Xu, and B.-L. Lu, ``Transfer learning for {EEG}-based brain-computer
  interfaces: {A} review of progress made since 2016,'' \emph{IEEE Trans. on
  Cognitive and Developmental Systems}, vol.~14, no.~1, pp. 4--19, 2020.

\bibitem{drwuNTL2023}
W.~Zhang, L.~Deng, L.~Zhang, and D.~Wu, ``A survey on negative transfer,''
  \emph{IEEE/CAA Journal of Automatica Sinica}, vol.~10, no.~2, pp. 305--329,
  2023.

\bibitem{li2021can}
W.~Li, W.~Huan, B.~Hou, Y.~Tian, Z.~Zhang, and A.~Song, ``Can emotion be
  transferred?---{A} review on transfer learning for {EEG}-based emotion
  recognition,'' \emph{IEEE Trans. on Cognitive and Developmental Systems},
  vol.~14, no.~3, pp. 833--846, 2021.

\bibitem{zhou2019transferable}
H.~Zhou and K.~Chen, ``Transferable positive/negative speech emotion
  recognition via class-wise adversarial domain adaptation,'' in \emph{Proc.
  IEEE Int'l Conf. on Acoustics, Speech and Signal Processing}, Brighton, UK,
  May 2019, pp. 3732--3736.

\bibitem{kaya2017video}
H.~Kaya, F.~G{\"u}rp{\i}nar, and A.~A. Salah, ``Video-based emotion recognition
  in the wild using deep transfer learning and score fusion,'' \emph{Image and
  Vision Computing}, vol.~65, pp. 66--75, 2017.

\bibitem{ngo2019facial}
T.~Q. Ngo and S.~Yoon, ``Facial expression recognition on static images,'' in
  \emph{Proc. Future Data and Security Engineering}, Nha Trang City, Vietnam,
  Nov. 2019, pp. 640--647.

\bibitem{sugianto2019cross}
N.~Sugianto and D.~Tjondronegoro, ``Cross-domain knowledge transfer for
  incremental deep learning in facial expression recognition,'' in \emph{Proc.
  Int'l Conf. on Robot Intelligence Technology and Applications}, Daejeon,
  South Korea, Nov. 2019, pp. 205--209.

\bibitem{zhao2020speech}
H.~Zhao, N.~Ye, and R.~Wang, ``Speech emotion recognition based on hierarchical
  attributes using feature nets,'' \emph{Int'l Journal of Parallel, Emergent
  and Distributed Systems}, vol.~35, no.~3, pp. 354--364, 2020.

\bibitem{park2019toward}
S.~Park, J.~Kim, J.~Jeon, H.~Park, and A.~Oh, ``Toward dimensional emotion
  detection from categorical emotion annotations,'' \emph{arXiv:1911.02499},
  2019.

\bibitem{Mohammad2018}
S.~M. Mohammad, ``Obtaining reliable human ratings of {Valence}, {Arousal}, and
  {Dominance} for 20,000 {English} words,'' in \emph{Proc. Annual Conf. of the
  Association for Computational Linguistics}, Melbourne, Australia, Jul. 2018.

\bibitem{bradley1999affective}
M.~M. Bradley and P.~J. Lang, ``Affective norms for {E}nglish words ({ANEW}):
  {I}nstruction manual and affective ratings,'' The Center for Research in
  Psychophysiology, University of Florida, Tech. Rep., 1999.

\bibitem{zhou2020fine}
F.~Zhou, S.~Kong, C.~C. Fowlkes, T.~Chen, and B.~Lei, ``Fine-grained facial
  expression analysis using dimensional emotion model,'' \emph{Neurocomputing},
  vol. 392, pp. 38--49, 2020.

\bibitem{warriner2013norms}
A.~B. Warriner, V.~Kuperman, and M.~Brysbaert, ``Norms of valence, arousal, and
  dominance for 13,915 {E}nglish lemmas,'' \emph{Behavior Research Methods},
  vol.~45, pp. 1191--1207, 2013.

\bibitem{Shannon1948}
C.~E. Shannon, ``A mathematical theory of communication,'' \emph{The Bell
  System Technical Journal}, vol.~27, no.~3, pp. 379--423, 1948.

\bibitem{settles2008analysis}
B.~Settles and M.~Craven, ``An analysis of active learning strategies for
  sequence labeling tasks,'' in \emph{Proc. Conf. on Empirical Methods in
  Natural Language Processing}, Honolulu, HI, Oct. 2008, pp. 1070--1079.

\bibitem{culotta2005reducing}
A.~Culotta and A.~McCallum, ``Reducing labeling effort for structured
  prediction tasks,'' in \emph{Proc. AAAI Conf. on Artificial Intelligence},
  vol.~5, Pittsburgh, PA, Jul. 2005, pp. 746--751.

\bibitem{pan2010domain}
S.~J. Pan, I.~W. Tsang, J.~T. Kwok, and Q.~Yang, ``Domain adaptation via
  transfer component analysis,'' \emph{IEEE Trans. on Neural Networks},
  vol.~22, no.~2, pp. 199--210, 2010.

\bibitem{wang2017balanced}
J.~Wang, Y.~Chen, S.~Hao, W.~Feng, and Z.~Shen, ``Balanced distribution
  adaptation for transfer learning,'' in \emph{Proc. IEEE Int'l Conf. on Data
  Mining}, New Orleans, LA, November 2017, pp. 1129--1134.

\bibitem{busso2008iemocap}
C.~Busso, M.~Bulut, C.-C. Lee, A.~Kazemzadeh, E.~Mower, J.~N. Kim,
  Samueland~Chang, S.~Lee, and S.~S. Narayanan, ``{IEMOCAP}: {I}nteractive
  emotional dyadic motion capture database,'' \emph{Language Resources and
  Evaluation}, vol.~42, no.~4, pp. 335--359, 2008.

\bibitem{poria2019meld}
S.~Poria, D.~Hazarika, N.~Majumder, G.~Naik, E.~Cambria, and R.~Mihalcea,
  ``{MELD}: {A} multimodal multi-party dataset for emotion recognition in
  conversations,'' in \emph{Proc. 57th Annual Meeting of the Association for
  Computational Linguistics}, Florence, Italy, Jul. 2019, pp. 527--536.

\bibitem{grimm2008vera}
M.~Grimm, K.~Kroschel, and S.~Narayanan, ``The {Vera am Mittag German}
  audio-visual emotional speech database,'' in \emph{Proc. IEEE Int'l Conf. on
  Multimedia and Expo}, Hannover, Germany, Jun. 2008, pp. 865--868.

\bibitem{baevski2020wav2vec}
A.~Baevski, Y.~Zhou, A.~Mohamed, and M.~Auli, ``wav2vec 2.0: {A} framework for
  self-supervised learning of speech representations,'' in \emph{Proc. Int'l
  Conf. on Neural Information Processing Systems}, vol.~33, Virtual Event, Dec.
  2020, pp. 12\,449--12\,460.

\bibitem{panayotov2015librispeech}
V.~Panayotov, G.~Chen, D.~Povey, and S.~Khudanpur, ``Librispeech: {A}n {ASR}
  corpus based on public domain audio books,'' in \emph{Proc. Int'l Conf. on
  Acoustics, Speech and Signal Processing}, South Brisbane, Australia, Apr.
  2015, pp. 5206--5210.

\bibitem{borgwardt2006integrating}
K.~M. Borgwardt, A.~Gretton, M.~J. Rasch, H.-P. Kriegel, B.~Sch{\"o}lkopf, and
  A.~J. Smola, ``Integrating structured biological data by kernel maximum mean
  discrepancy,'' \emph{Bioinformatics}, vol.~22, no.~14, pp. 49--57, 2006.

\bibitem{holm1979simple}
S.~Holm, ``A simple sequentially rejective multiple test procedure,''
  \emph{Scandinavian Journal of Statistics}, vol.~6, no.~2, pp. 65--70, 1979.

\end{thebibliography}


\end{document}